# Qualitative Event Perception: Leveraging Spatiotemporal Episodic Memory for Learning Combat in a Strategy Game

**Will Hancock** WWHANCOCK@U.NORTHWESTERN.EDU
**Kenneth D. Forbus** FORBUS@NORTHWESTERN.EDU
Qualitative Reasoning Group, Northwestern University, Evanston, IL

## Abstract

Event perception refers to people's ability to carve up continuous experience into meaningful discrete events. We speak of finishing our morning coffee, mowing the lawn, leaving work, etc. as singular occurrences that are localized in time and space. In this work, we analyze how spatiotemporal representations can be used to automatically segment continuous experience into structured episodes, and how these descriptions can be used for analogical learning. These representations are based on Hayes' notion of histories and build upon existing work on qualitative episodic memory. Our agent automatically generates event descriptions of military battles in a strategy game and improves its gameplay by learning from this experience. Episodes are segmented based on changing properties in the world and we show evidence that they facilitate learning because they capture event descriptions at a useful spatiotemporal grain size. This is evaluated through our agent's performance in the game. We also show empirical evidence that the perception of spatial extent of episodes affects both their temporal duration as well as the number of overall cases generated.

## 1. Introduction

People carve up the continuous world into discrete events. We talk about drinking a cup of coffee in the morning, emptying the trash, filling the dishwasher, etc. These events are localized, i.e. spatially bounded and temporally extended. In psychology research, the ongoing segmentation of continuous experience is known as *event perception (*Zacks, 2020*)*.

In qualitative reasoning research, Hayes proposed the notion of *histories* (Hayes, 1989;1978) as a general framework for representing change in continuous domains. Whereas the situation calculus describes changes in terms of discrete events and provides no spatial constraints (leading to the infamous Frame Problem), histories represent change in terms of pieces of space-time, based on the objects involved in the changes.

In this paper, we show how history-based representations can be used to automatically carve up continuous experience into discrete episodes. We investigate the relationship between representational choices and temporal segmentation, and then show how this affects an agent's ability to learn from past experience. Our task is analogically learning qualitative states of defense for cities in the strategy game *Freeciv*[1], where our agent records traces of military battles and makes decisions in the game based on knowledge generalized from these experiences. Our representations build upon existing work on Qualitative Spatiotemporal Episodic Memory (QSTEM) (Hancock and Forbus, 2021) that developed a qualitative representation of spatiotemporally localized change. In this work, our experimental results show that these representations facilitate learning because they capture events at a useful grain size. Furthermore, we analyze the relationship between representational assumptions and episodic segmentation, showing how extended representations of space affect temporal segmentation. Specifically, the kinds of spatial representations used affect both individuation, i.e. the number of episode instances, as well as the quality of the descriptions that go into those instances. All else being equal, a larger perceived battlefield means more participants which in turn means a longer battle. If an agent's view of a battle is the entire world, then fewer episodes will be generated, and the descriptions will lack parsimony. Similarly, battles that are perceived too locally result in more episodes, but the contained knowledge is less useful for learning. Whereas it is commonly assumed that there is a monotonic relationship between the number of examples available to learn and the ability to learn, we find that on the contrary, event perception is important because fewer but higher quality examples can allow better learning performance than that with more, lower quality examples.

[1] https://www.freeciv.org/

This work focuses on a single event type, military battles, in a single domain. In general, the proposed model applies to a wide range of event types and domains, i.e. complex quantity conditioned events that involve social gatherings, classrooms, liquids, etc. Locality is a powerful constraint on event individuation, which is one of the reasons that histories are a useful representation mechanism; an agent's ability to segment events, and therefore learn, depends on its conception of space. We are currently evaluating these same representations for learning commonsense events in a cognitive robotics environment called AILEEN (Mohan et. Al, 2021). We begin by discussing QSTEM, our analogical learning stack, and our testbed domain, Freeciv.

## 2. Background & Motivation

### 2.1 Qualitative Spatiotemporal Episode Memory

Our qualitative representation (QSTEM) is based on Hayes' notion of histories, i.e. bounded pieces of spacetime. Histories were developed around intuitive models of human thinking about space and time. They emerged in part as an ontological solution to difficulties in representing continuous entities. A canonical example involves liquids; pieces of lakes are not readily individuated, so it is more natural to reify them based on spatial location. Histories have been used extensively in qualitative reasoning research (Forbus, 2019) and so we hypothesize that they will be useful for learning with episodic memory.

One component of QSTEM is a qualitative representation of change. These descriptions reify changing attributes of spatially extended entities as qualitative temporal intervals. For example, we can describe the period of decreasing height of a falling object, of the buildup of troops in preparation for a siege, of the increase of a player's gold, etc.

Here we aim to show how history-based representations can be used for event segmentation, and how the resulting episodes can be used for learning. The automatic spatial and temporal adaptation based on the entities that make up the descriptions means that these descriptions rely on general purpose encoding schemes, allowing the agent to adapt to new situations more easily.

## 2.2 Analogical Learning

One goal of this research is to demonstrate how analogy can support spatiotemporal learning. To do this, we use the Structure Mapping Engine (SME) (Forbus et al. 2017) for comparison, the MAC/FAC retrieval system (Forbus et al., 1995), and the SAGE generalization system (McLure et al. 2015). These analogical mechanisms are tightly integrated in the underlying reasoning engine and provide the mechanisms for retrieval, comparison, and transfer. There is a growing body of evidence that human cognition and therefore commonsense reasoning depends on structured representations of the world. The perceptual world around us is highly structured, both temporally and spatially, and so we think that analogical processes play an important role in learning.

### *2.2.1 SME*

The Structure Mapping Engine (SME) (Forbus et al., 2017) is a computational model of matching in Structure Mapping Theory (Gentner, 1983). SME provides a mechanism for determining similarity between two structured descriptions. It does this by generating a set of mappings between a base and target. A mapping aligns entities and expressions between the descriptions. A structural similarity score is computed for each mapping that prefers higher-order structure and entities that participate in systems of relations. SME additionally computes a set of candidate inferences, or surmises about the target made on the basis of common structure plus the base representation.

### *2.2.2 MAC/FAC*

Analogical retrieval is performed by MAC/FAC (Forbus et al., 1995), which stands for “Many are Called, Few are Chosen” because it uses two stages of map/reduce for scalability. The inputs consist of a probe case and a case library. The MAC stage, in parallel, computes dot products over vectors that are automatically constructed from structured descriptions, such that each predicate, attribute, and logical function are dimensions in the vector and whose magnitude in each dimension reflects their relative prevalence in the original structured description. The best mapping, and up to two others, are passed to the FAC stage. FAC compares the best structured descriptions from the MAC stage to the input probe using SME. The best match plus up to two others, is then returned. The inexpensive nature of the MAC stage allows for scalability, while the FAC stage allows for sensitivity to structural similarity.

### *2.2.3 SAGE*

SAGE, or the Sequential Analogical Generalization Engine (McLure et al., 2015) is an analogical generalizer. It utilizes MAC/FAC for retrieval, and therefore SME at its core. SAGE operates over *generalization pools* (aka gpools), each of which represents a concept. Each gpool consists of a set of generalizations and outlier examples. Given a new example and a gpool, MAC/FAC first finds the most similar generalization or outlier to the example. This retrieval results in a SME mapping, which aids in assimilation. Examples are only merged if their similarity score exceeds an assimilation threshold. Generalizations accumulate statistics over expressions and filter out low frequency expressions based on a probability cutoff. Unlike many ML algorithms, SAGE is an incremental learner, which eliminates the need for retraining a model. It is similar to k-means with outliers, but unlike k-means uses a human-normed similarity metric operating on structured representations. In addition, the number of generalizations falls out automatically (i.e. it does not need to be prespecified), allowing the flexible learning of disjunctive concepts. As part of the generalization procedure, SAGE lifts mapped entities, and replaces them with *generalized entities*, (GenEnts), a form of Skolem constant. GenEnts that correspond to quantitative arguments store statistics on the participating quantities. Specifically, cardinality, minimum, maximum, mean, and sum of squared error are collected. A generalization can be queried so that it returns these statistics for a specific quantitative generalized entity.

### 2.3 Freeciv

Freeciv is a complex turn-based strategy game based on the Civilization franchise. It involves playing as an emerging nation, vying against potentially hostile nations to either conquer the world or build a spaceship and found a new colony in space. Successfully achieving either outcome requires a player to manage their scientific progress, military, and economic growth all while conducting diplomatic relations.

Freeciv is an excellent domain for AI research due to its complexity. A typical game board consists of 4,000 tiles with varying terrain. Games typically last for hundreds of turns, and each turn involves many decisions. Some decisions are global across the entire civilization, such as setting the tax rate, determining the next technology to research, and engaging in diplomacy. Workers must be kept busy modifying terrain. Military units must defend cities and conduct attacks on opponents when at war. By contrast, Go is played on a 19x19 grid with uniform, immutable spatial properties which are always visible from the start. In addition, each turn in Go only involves a decision to place one piece.

Freeciv is especially useful for exploring histories because important behaviors happen at multiple grain sizes. Combat often evolves as the game progresses, where early engagements are typically quick and involve fewer participants, and later battles (and sieges) are more drawn out and involve more unit types. Describing military battles using histories should benefit learning because case descriptions adapt automatically based on entities that participate in the event.

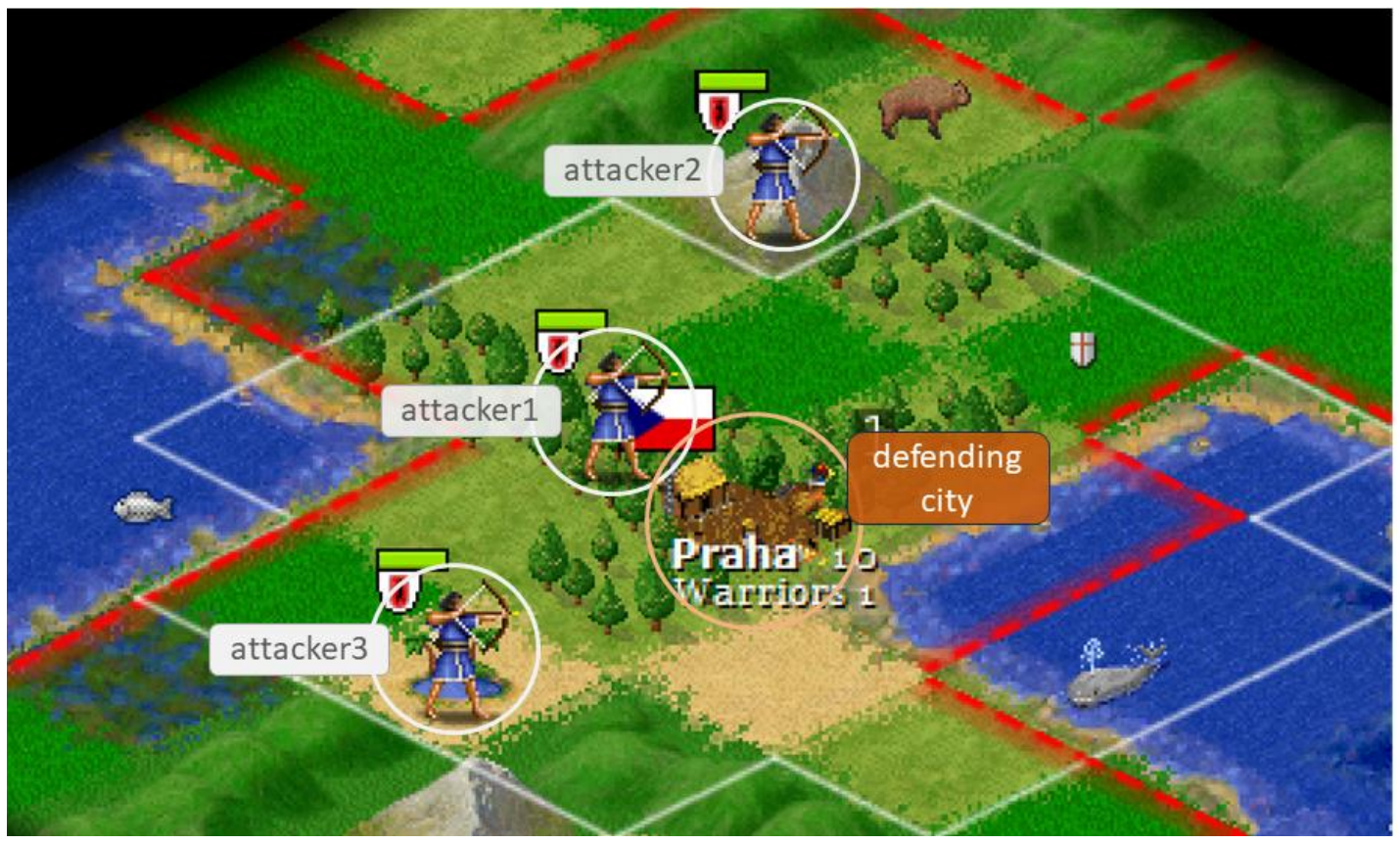

Figure 1: A battle consisting of three enemy units attacking a player's city.

## 3. Learning from Battles in Freeciv

We begin by motivating the relationship between event perception, i.e. determining the spatiotemporal bounds of an event, and learning from the resulting descriptions. Consider the scenario in Figure 1, where a group of opposing units are attacking a city. The hypothesis is that without histories, learning will be harder because localized events like battles cannot be adequately characterized. Specifically, spatial extension of an event directly affects temporal segmentation and thus event individuation. In Figure 1, assume that the archers (attacker1) nearest to the city (Praha) attack the city, that there is at least one defender inside Praha, and the archers (attacker1) lose the attack and are removed from the game. In Figure 2, this scenario is illustrated without extended spatial representations.

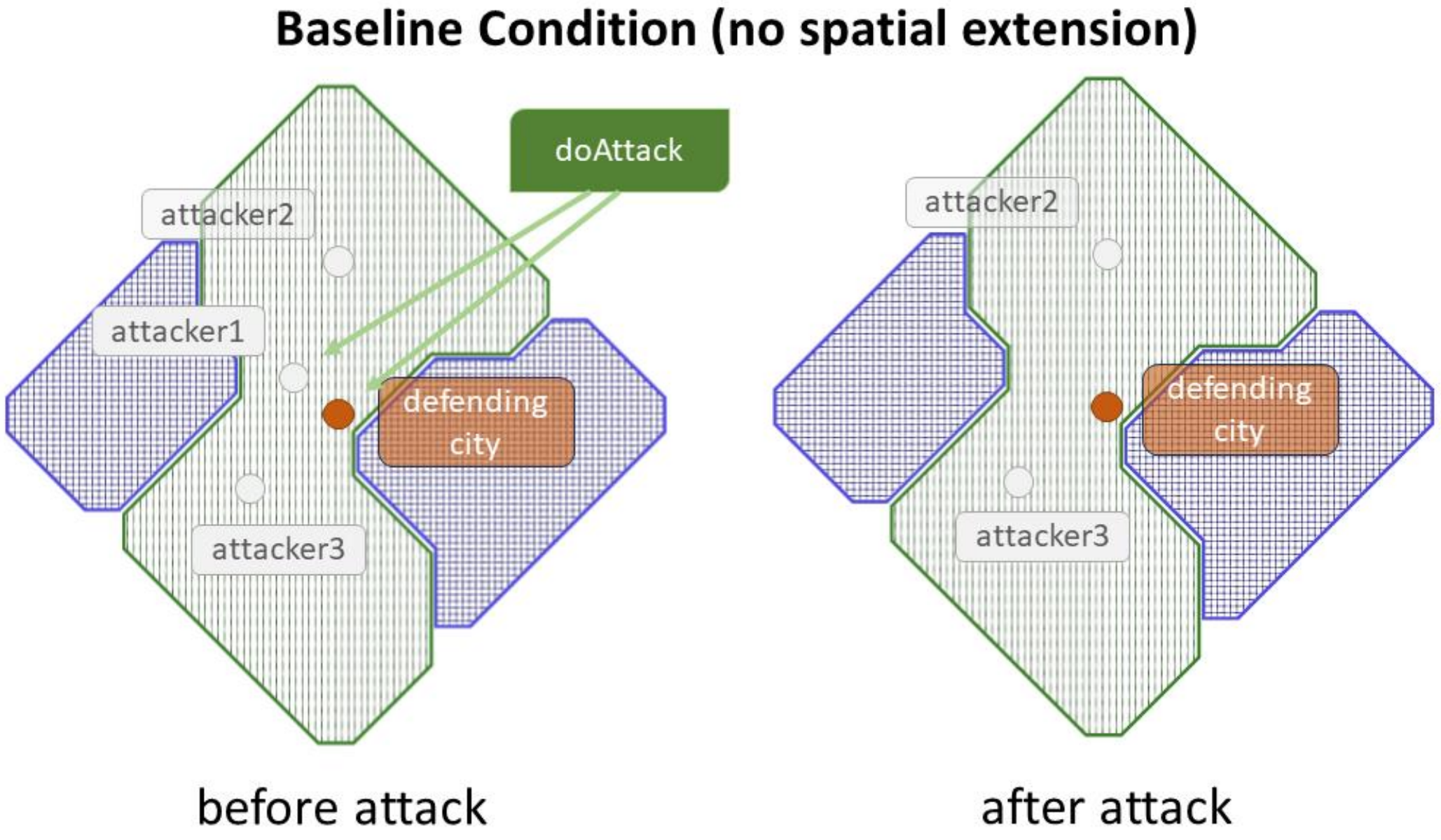

Figure 2: Before and after representation of attack action without extended spatial representations. Assume that attacker1 loses an attack on the defending city and is removed from the game. In this situation, the battle from the perspective of the city is over because there are no longer any spatially local enemy units, i.e. at the previous location of attacker1 or the defending city.

We can define quantity conditioned events (e.g. battles) in terms of perceptual entities (the event participants). That is, a battle is over when one side has been eliminated (its cardinality is zero). This definition then provides a notion of persistence for the event, i.e. the temporal extension lasts as long as two local opposing forces exist within some spatial context. Note the dependence on locality. For this work, we compare a *baseline* experimental condition that does not use extended spatial representations against the *histories* condition that does. For the baseline condition, the venue for the battle is represented by two locations (space in Freeciv is represented tiles), the coordinates of the attacker and defender. In Freeciv, the attack action primitive takes the form (doAttack ?attacker ?defender) where both <?attacker> and <?defender> are singular entities.

For Figure 1, these locations are the coordinates of attacker1 and the city Praha. After the attack, since the city is the only remaining entity at either location, the battle is over for the baseline condition. On the other hand, Figure 3 depicts the same situation given extended spatial representations. Here, the battle persists after attacker1 has been removed. To determine this, a notion of persistence for extended representations must be given. A natural way to construe this using qualitative representations is with a binary form of the *Region Connection Calculus* (RCC2) (Cohn et al., 1997). This calculus defines binary relations over spatial regions, where rcc2-connected relates regions that share at least one point. In the above example, the battle persists because there are still opposing unit participants after the initial attack, i.e. the relation (rcc2-connected attacker3 city) holds.

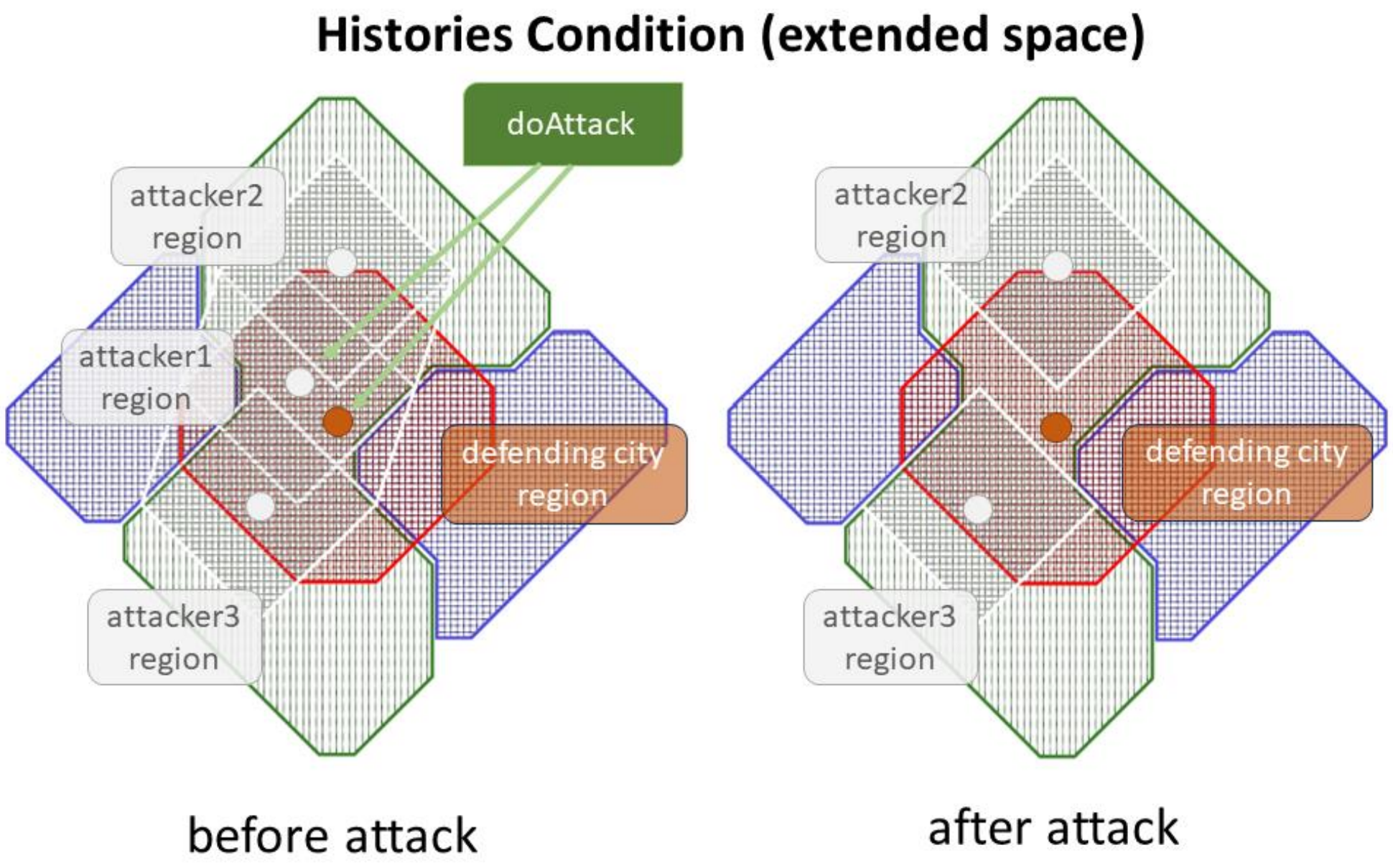

Figure 3: The same scenario as above is depicted, this time with extended spatial representations. Attacker1 loses its assault against the city and is removed from the game. However, there remain enemy units that are local to the city because their footprints are connected. In this case, the battle persists until attacker2 and 3 successfully take the city or flee.

Given an understanding of how spatial representations affect the temporal extent of episodes, we next focus on how this impacts learning. Consider the question of how many military units are needed to defend a city. The goal for our agent is to learn a model that differentiates between adequate and inadequate defensive states. Assume that the scenario depicted in Figure 1 eventually leads to the city being conquered. In the histories condition (Figure 3), the battlefield encompasses the city and three attacking units, which is determined by connectivity (rcc2-connected) of participant footprints. Here, the battle persists until the city is taken (assuming the locality conditions hold), resulting in a single episode containing all three attackers as participants. On the other hand, for the baseline condition, the same scenario results in three episodes, corresponding to each discrete attack event. Each describes a battle with one attacker, i.e. the spatiotemporal bounds are more fine-grained. The outcomes of each episode are also different. The first two result in successes for the city, while the third results in a failure. This has implications for learning and reasoning, namely that the baseline condition does not capture enough salient knowledge about the event. The cases are too localized and therefore result in an agent reasoning about situations with single attackers. An agent could avoid using these kinds of representations by simply reasoning about all entities that exist in a scenario, i.e. all visible units throughout the world map. This forgoes localized space by simply considering all space. In this case, if local spatial representations are unavailable, then battles cannot be distinguished from wars, because battles are inherently local. In the next section, we describe how these cases are constructed automatically.

### *3.1.1 Case Construction*

Case construction for this experiment generates a structured description of military battles. Generated cases are based on QSTEM, with two additions. First, QSTEM segmented time solely on the basis of qualitative representations of change, i.e. intrinsic properties of quantities, whereas in this work, episodes can be defined as events which can have more complex start and end conditions, e.g. battles. Second, this work reifies and learns statistics over quantities that are included in cases, and later we describe how these statistics are used for decision making.

We begin by describing how we define military battles, which are constructed from the viewpoint of a player's cities. Battles start when an enemy attacks a city, and end when either the city has been conquered or there are no more local enemy units. Figure 3 shows a battle with enemy units (e.g. attacker2) and a defending city region, which are all *compound spatial entities*. We call these entities footprints (Hancock et al., 2020), which are spatial regions that are constructed out of more primitive regions. For this work, we use three kinds of footprints:

1. City footprints: Defined by the twenty-one tiles in a city's region
2. Unit footprints: The region bounded by a unit's max movement in one turn
3. Unit group footprints: The region defined by a set of spatially local unit footprints

All footprints are constructed by computing the convex hull over constituent entities. During an ongoing battle, participants consist of the defending city footprint and any spatially local (i.e. rcc2-connected) unit group footprint.

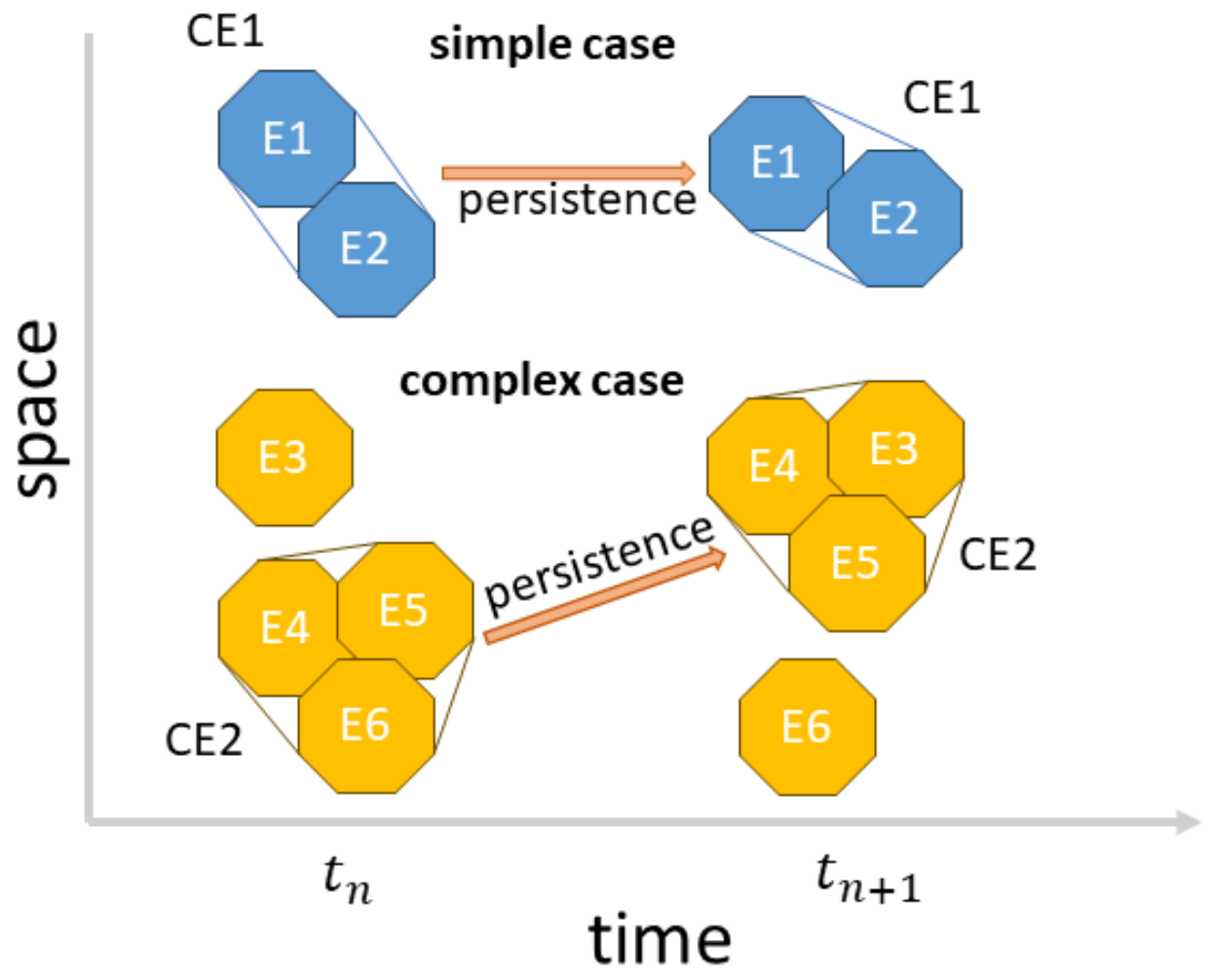

Figure 4: Change for compound entities (CE) over time for a simple and complex case. In the simple case, constituent members of CE1 remain the same from $t_n$ to $t_{n+1}$, so the entity CE1 persists. In the complex case, merging and splitting occurs across subsequent timepoints. In this work, the larger entity persists (CE2), while e3 is removed, and e6 appears at time $t_{n+1}$.

### *3.1.2 Persistence for Compound Entities*

Footprints are reified so that their attributes can be described. For military battles, one might describe the size of opposing forces (i.e. an aggregate quantity over relevant constituent units). We could also describe these quantities in terms of change, i.e. the size is increasing or decreasing. Hancock and Forbus (2021) showed that using qualitative descriptions of change helps an agent learn production decisions in Freeciv from a human. To reify these kinds of descriptions of change, we must first specify how compound entities persist. In other words, what makes an entity the same from one time to another if it has lost or gained (possibly both) some of its parts? For this work, persistence is determined by set intersection of constituent entities of a footprint. This depends on being able to individuate those entities (i.e. they can be named)[2].

Figure 4 shows an example of persistence for two cases: a simple and complex case. In the simple case, the compound entity CE1 remains the same from $t_n$ to $t_{n+1}$ whereas in the complex case, the entity gains and loses parts. We still have not described how persistence is determined, e.g. in the complex case, CE2 at $t_n$could persist as either of the two entities at $t_{n+1}$. For this work, we use a simple heuristic which states that the entity persists as the largest entity. Given this definition, we can now describe how compound entities change over time, e.g. CE2 remains the same size. In the next section, we describe how properties are recorded for entities over time, resulting in a series of fluents. These fluents are then used to construct qualitative descriptions of change.

### *3.1.3 Recording Fluents*

Knowledge about entities and their changing attributes is recorded at each timestep. In this work, time is indexed by significant events that occur in the game. Examples are the start of a new Freeciv turn, when entities are created and destroyed, when units move, when resources are produced, etc.

We declaratively define a set of quantities that are recorded. Figure 5 shows one definition for the number of enemy units local to a player's city. Each definition is a triple, where the second argument names the quantity and the third argument specifies the intension of the set, i.e. the set of entities that the quantity should be recorded for. This work uses the same quantities defined in (Hancock and Forbus, 2020) as well as Figure 5, which is new. These definitions are used to record a set of fluents at each timestep. For each quantity type, first, the set extension is determined. From Figure 5, the form (GenericInstanceFn … ) is queried in the current Freeciv context to return a set of relevant entities, e.g. (FC-City-Chicago FC-City-Boston). For each of these entities, an additional query is generated using the defined quantity type. This would be (((MeasurableQuantityFn enemyUnitCardinality) FC-City-Chicago), …). These forms are again queried against the current scenario context and call outsourced predicates that return the current

```
(quantityName BattleOpposingUnitCardinality
  ((MeasurableQuantityFn enemyUnitCardinality)
   (GenericInstanceFn
    (CollectionSubsetFn FreeCiv-City
      (TheSetOf ?city
        (and (sovereignAllegianceOfOrg ?city (IndexicalFn currentRole))
             (isa ?city FreeCiv-City)))))))
```

Figure 5: Quantity definition for the cardinality of opposing units local to a city.

[2] For more complex cases, e.g. liquids, persistence could be determined by qualitative connectivity (rcc2). We leave this for future work.

quantity value. For (((MeasurableQuantityFn enemyUnitCardinality) FC-City-Chicago), this might be (FreecivUnitCountFn 0) if there are no local enemy units at that instant. These values are stored as fluents in a map in memory as follows:

(<?quantity-name> . <?entity-name>): $((value_n\ t_n)\ (value_{n-1}\ t_{n-1})\ \ldots)$

At each update, each entry is checked to determine qualitative change. Each quantity type (e.g. BattleOpposingUnitsCardinality) is associated with a set of qualitative encodings. These associated are made declaratively in the form:

(quantityEncodingSchemeFor <?quantity-name> <?encoding-type>)

We use two previously defined encoding types from Hancock and Forbus (2021): the derivative sign encoding and magnitude sign encoding. For example, the BattleOpposingUnitsCardinality quantity is associated with both types. Assume that up until time $t_n$, Chicago has no local enemies, i.e. the set of recorded fluents is:

(BattleOpposingUnitsCardinality $Chicago$): $\left(\left(t_n\ (UnitCount\ 0)\right)\left(t_{n-1}\ (UnitCount\ 0)\right) \ldots\right)$

Next, assume that at $t_{n+1}$ a new quantity $(t_{n+1}(UnitCount\ 1))$ is sampled. For the magnitude sign encoding, this means that the encoding was previously *None*, while the new encoding is *Some*. Here, the existing temporal interval corresponding to the *None* encoding is closed, i.e. its temporal endpoint is set to the current time, and a new temporal interval corresponding to the *Some* encoding is instantiated. This checking for qualitative change occurs each time the set of quantities is sampled. In addition to recording quantitative fluents at each instant, the set of event participants is also updated. That is, a relation is reified (eventParticipant <?event-name> <?participant-name>). This knowledge is used for case construction as a record for determining which entity descriptions are included in the case, which we describe next.

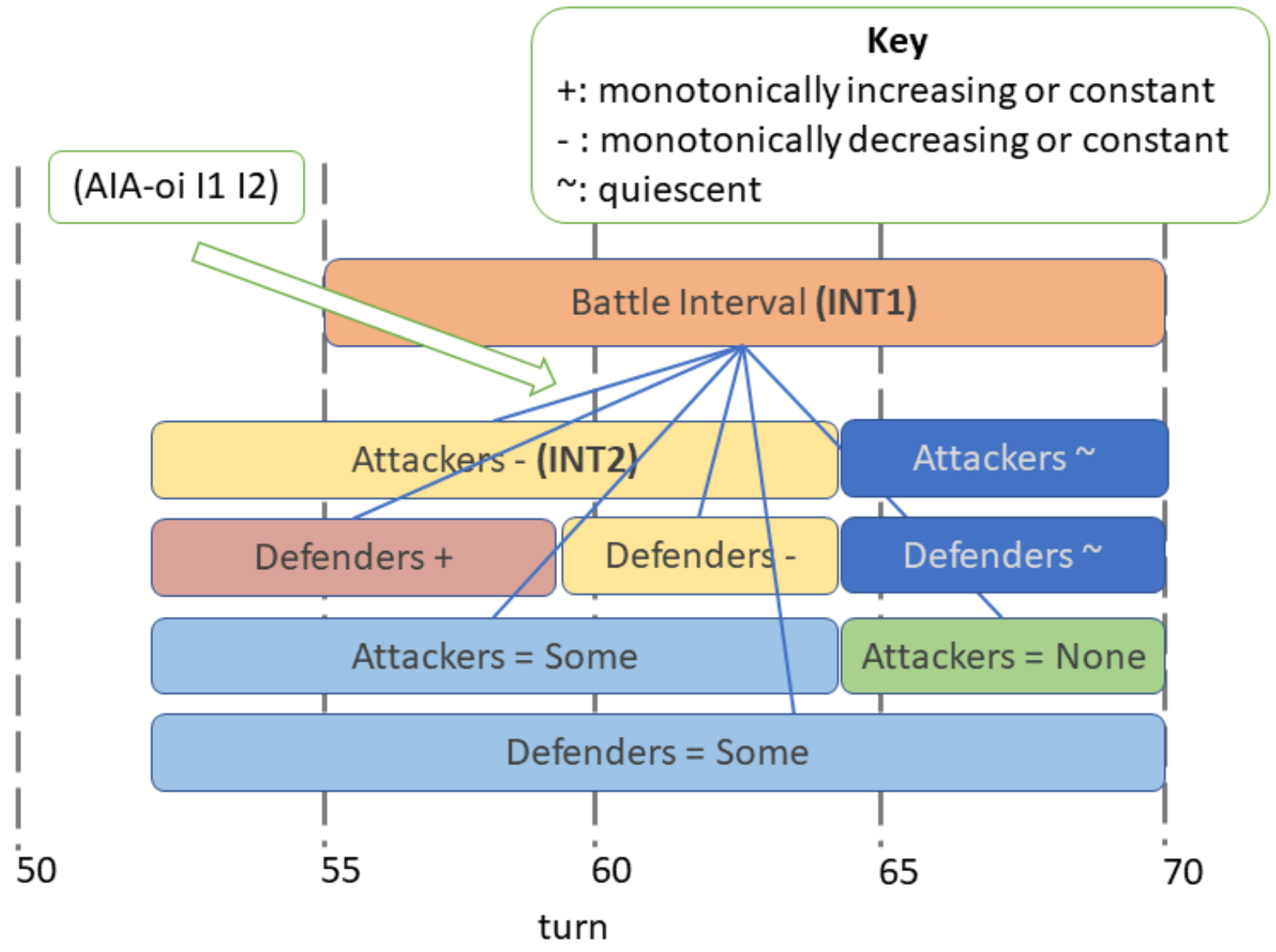

Figure 6: Qualitative temporal intervals for an example military battle. The interval I1 is related to all other temporally local intervals (i.e. not AIA-precedes or AIA-precededBy) using Allen's Interval Algebra, e.g. (aia-overlappedBy I1 I2).

The next step in case construction is to reify a set of histories for participant entities (including the reified event). This is determined by reasoning about the spatiotemporal bounds of the history of each constituent. An example of the history generated for the event is shown in Figure 6. The interval I1 represents the temporal extent of the battle. These intervals are reified and their attributes are described using Davidsonian relations, e.g. (monotonicallyNonIncreasingIn I2 (Attackers FC-City-Chicago)). This interval I1 is related to all other qualitative intervals using Allen's Interval Algebra (Allen, 1983). Histories for other event participants are generated as well in the same manner, i.e. the defending city footprint and other participant unit group footprints. The temporal extent for each history is determined by some qualitative temporal interval. For example, for unit group footprints, this is the derivative sign encoding of the size of the footprint. The history for a unit group footprint might then be a period of increase in size for the entity. These associations are made declaratively in the form:
(quantityEncodingForRegion <?qtype> <?encoding> <?spatial-entity-type>).
For unit group footprints, this is
(quantityEncodingForRegion regionSize DerivativeSignEncoding UnitGroupFootprint).
Type statements for all entities are added as well, e.g. (isa <?region> UnitGroupFootprint). Finally, quantities are reified at the start and end of the event. These statements take the form:

(holdsIn (StartFn E1) (valueOf (Attackers Region1) 2))

(holdsIn (EndFn E1) (valueOf (Attackers Region1) 0))

In English, these can be read as "there were two attackers present at the beginning of the battle, and none present at the end". These statements are generated for the start and end values of each quantity type, i.e. values at the temporal bounds of the event.

### *3.1.4 Learning a Model*

After each battle has ended, a case is constructed and added to a success or failure gpool, depending on the outcome. Learning is incremental, with cases being generalized in the order that they are experienced. The general process for learning a SAGE model proceeds as follows. Assume that an agent experiences a series of failures. For the first battle event, the case is added to the failure gpool as an outlier. For the next example, the case is compared against the existing outlier. If it is sufficiently similar to the first case (as determined by SME's similarity score being above the SAGE assimilation threshold), then the two cases are merged into a single generalization. Otherwise, the second case is also added as an outlier. Learning continues in this matter as each new case is generalized into the model. Each cluster of cases (generalizations) can be seen as a learned schema for some disjunctive conceptualization of the overall model. For the battle failure gpool, these clusters describe different states where battles have been lost, e.g. these might describe a city in the desert with no city walls and three defenders, and a city on a hill with no defenders.

As part of this learning process, statistics are accumulated for quantities that appear in the structured case descriptions. Recall that these quantities take the form (holdsIn <?time> (valueOf <?qtype> <?value>)). Here, <?value> is a typed numerical statement, e.g. (FreecivUnitCountFn 3). SAGE automatically accumulates statistical knowledge for typed quantities that appear in structured descriptions. This process proceeds as follows. Assume that a new case describing a battle has been constructed, and that this case is being assimilated into a SAGE generalization. Figure 7 shows an example mapping, i.e. corresponding statements, from the case to the Gpool generalization. As part of this generalization process, mapped terms are replaced by logical functions called generalized entities that denote the lifted type for ground entities (GenEntFn <?index>), here abbreviated as GEFn. These reified entities are associated with a history of their ground values, e.g. (UnitCount 2), (UnitCount 1), (UnitCount 0) from each case in the generalization (it is also possible that a case has no corresponding description). For purely numeric generalized entities, statistics are accumulated as each new quantity is seen (e.g. (2, 1, and 0)). Specifically, the cardinality, minimum, maximum, mean, and sum of squared error is computed. These statistics are used for decision making, as described next.

| **Case Description** | **Gpool Generalization** |
|---|---|
| (holdsIn (EventTimeFn 50) (valueOf (Attackers Region1) (UnitCount 3))) | (holdsIn (GEFn 0) (valueOf (Attackers (GEFn 5)) (UnitCount (GEFn 1))) |
| (isa STEntity1 DefensiveEpisodeFailure) | (isa (GEFn 2) DefensiveEpisodeFailure) |
| (constantIn I1 (Attackers Region1)) | (constantIn (GEFn 3) (Attackers GEFn 4)) |

Figure 7: Example statements from an incoming case (left) and the corresponding SAGE generalization that it has mapped to. Here, *GEFn* is an abbreviation for *GeneralizedEntityFn*, which is a logical function that denotes a generalized entity. In this example, (GEFn 1) denotes an entity that is associated with a set of quantities. SAGE automatically accumulates statistics over the distribution defined by these quantities, and the system uses these statistics for decision making.

### *3.1.5 Decision Making*

Decision making is driven by analogical comparison of the agent's current situation to past failures. The idea is to determine if the current spatiotemporal context for a city is similar to one in which a city lost the battle. If this is the case, then a decision is made to improve the city's military system. The agent divides games into phases based on qualitative properties. It starts out in a growth phase, where the number of cities is monotonically nondecreasing. The conquest phase starts when the number of invaders becomes greater than zero. Decision making for this experiment starts in the conquest phase, which allows time to grow one's empire before focusing on defense.

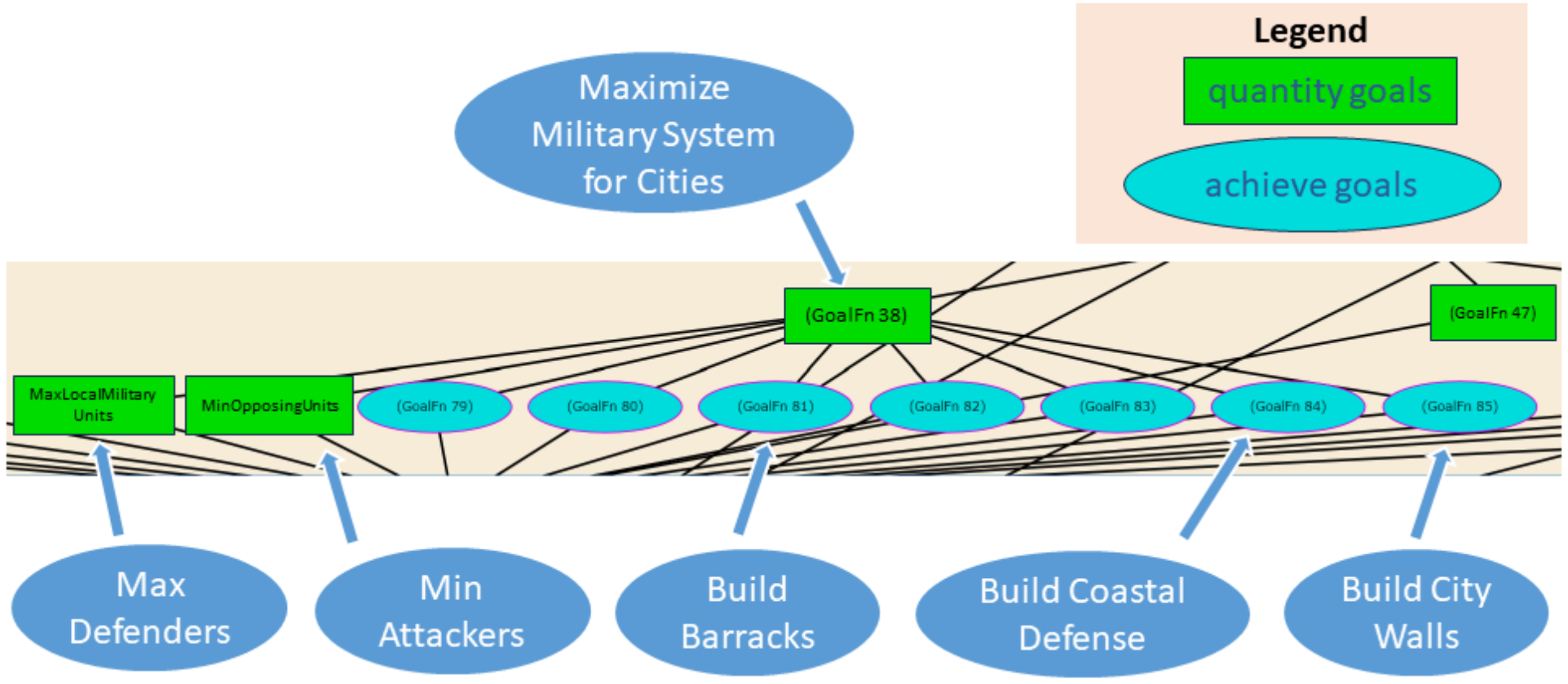

Figure 8: A subgraph of the goal network depicting maximization of city military systems (parent) and its influences.

The overall idea is for an agent to improve its gameplay by learning a domain model for defense. To assist with this decision making, we use an existing qualitative model of Freeciv (Forbus and Hinrichs, 2019; Hinrichs and Forbus, 2012). This model describes influences between quantity types in the game. These include measurable quantity types, e.g. the number of units that a player has, as well as abstract quantity types, e.g. the strength of a player's military system. For this experiment, decision making leverages the abstract goal to maximize a city's military system, depicted in Figure 8 along with its influences.

Each node in the goal network has a set of influences in the form of achieve and quantity influence goals. These can be seen as a set of subgoals, e.g. building city walls and maximizing the number of military units (Figure 8). Here, the *Max Defenders* and *Min Attackers* goals have been manually generated and added as subgoals for the city military system node, providing causal knowledge for improving decision-making.

For this experiment, determining the qualitative state of a city's military system depends on comparing its current state to previous battle failures. After a player's civilization has passed its growth phase, actions are determined through analogical retrieval. First, a case is constructed for the city that is making a decision, hereafter referred to as the probe case. Recall that for learning a model, cases are generated at the end of each battle and generalized in a SAGE gpool. For case generation at decision time, the city is not necessarily engaged in battle, and if it is, its temporal end has not been determined. In this sense, knowledge in the probe case is incomplete with respect to descriptions of full battles. However, a comparison can still be made by reifying any known histories, e.g. local enemy unit groups and relying on structural similarity between partial event descriptions. Recall that for this work, descriptions of battles consist of histories for the reified event (shown in Figure 6) as well as each participant entity. For the probe case, histories for all event participants are reified, but not the reified event. Participants are determined at the time of decision making, i.e. all spatially local unit group footprints as well as the city footprint are included in the case. For the probe case, quantities are sampled at the time of case construction and represented as:

(holdsIn (StartFn <?event-skolem-fn>) (valueOf <?quantity-type> <?quantity-value>)).

The idea is that an analogical retrieval for the probe case construes current quantity values as the hypothetical start of the event. In other words, the agent makes an analogical comparison if a battle were to break out at that timepoint.

For decision making, first, a probe case is constructed for a city. Next, an analogical retrieval is performed, comparing this case against both failure and success gpools. If no mapping is returned, or the retrieval maps to the success gpool (i.e. the agent predicts that the city in its current state will win a battle), then a default decision-making procedure[3] is invoked, which is shared across experimental conditions. If the failure gpool is mapped, then the next step is to determine an appropriate decision for improving the city's defensive state, using influences on the city military system goal as mentioned above. The agent uses this information to determine which subgoals are not currently met. Recall that there are two kinds of goals in the goal network: achieve and quantity constraint goals. Determining lack of satisfaction for achieve goals is straightforward, e.g. for the goal of building city walls, does it have them or not? Determining lack of satisfaction for quantity constraint goals is more complex and requires learning *limit points*, or significant points in a quantity space where conditions on either side of the point are qualitatively distinct (in this case sufficient/insufficient for defense). Using histories to learn limit points grounds quantity constraint goals into a specific spatiotemporal context. For example, one such limit point may be defined as the number of local military units that a city should station to successfully defend itself.

[3] This procedure was introduced in Hinrichs and Forbus (2016) and leverages the qualitative model to make investment decisions about continuous quantities (e.g. should I build a marketplace to increase gold production).

To learn limit points like this, the statistical knowledge that is accumulated in SAGE generalizations is used. There are three main steps:

1. The computed mapping is queried to align current quantities with those from prior experience.
2. Statistics associated with prior quantities are used to determine possible failure.
3. In the case of failure, an action is proposed that would influence the current quantity.

For step one, the analogical mapping is queried to align current quantity values with those from prior experience. Consider the goal to maximize the number of military units. The mapping for this quantity might appear like so:

(h $t_n$ (valueOf milUnits (UnitCount 1)) → (h (GEFn 0) (valueOf milUnits (UnitCount (GEFn 1))). Here, (GEFn 1) denotes the generalized entity for values of this quantity. Assume that the historical values for this entity are (1 0 1). This means that previously, battle failures have been experienced when there has been one, zero, and one local military unit respectively. Step two in the decision-making process is to use these values to determine goal failure, depending on the goal direction, i.e. maximization or minimization. For maximization goals, the failure condition is: (current-quantity <= max(previous-quantities)). For minimization goals, the failure condition is (current-quantity >= min(previous-quantities)). For example, the current number of defenders is 1, the previously seen quantities are (1 0 1), and this quantity is associated with a maximization goal, so a failure is signaled. The third and final step is (given a determined failure) to propose an action that would influence the quantity in the desired direction. Here, the new limit point is not explicitly reified; rather, our agent queries the current execution context for possible actions that the city can take to properly influence the failed quantity type, e.g. building a new military unit would positively influence the number of defenders. This query depends on existing knowledge from the qualitative model that ties ground actions to quantity goals, which is described in more detail in Hinrichs and Forbus, 2016).

Given the military system goal node (Figure 8), a set of actions is proposed for each unsatisfied subgoal. Actions that are not possible are filtered out, e.g. one cannot build a SAM battery before the scientific advance of rocketry has been discovered. Out of the remaining subset, an action is chosen at random and executed for the city.

## 4. Experiment

In this experiment, the two conditions are *baseline* and *histories*. For the baseline condition, the agent learns from combat experience without histories, i.e. it does not leverage extended representations of space, which constrains how the resulting events persist. The histories condition leverages extended spatial representations (i.e. regions). Our main hypothesis is that gameplay performance will improve more for the histories condition because histories should be useful for event perception, i.e. they help scope events. We show this in two ways. The first is by comparing the agent's performance in the two conditions. Second, we examine the relationship between episodic learning and history-based representations by analyzing the episodes generated by each condition. We evaluate our agent's performance in the game using two metrics: overall civilization size measured by the number of cities, and the amount of gold in the treasury. Determining success in Freeciv is complicated; there are many factors that indicate success in the game. In general, a larger empire indicates overall success, but it is possible to focus solely on expansion while neglecting other aspects of one's civilization, and so we report on economic health as well. Both experimental conditions progress in a similar manner until around turn sixty when enemy invasions begin. Here, the histories condition has learned a better model of defense and so is better able to maintain a higher number of cities and its economy.

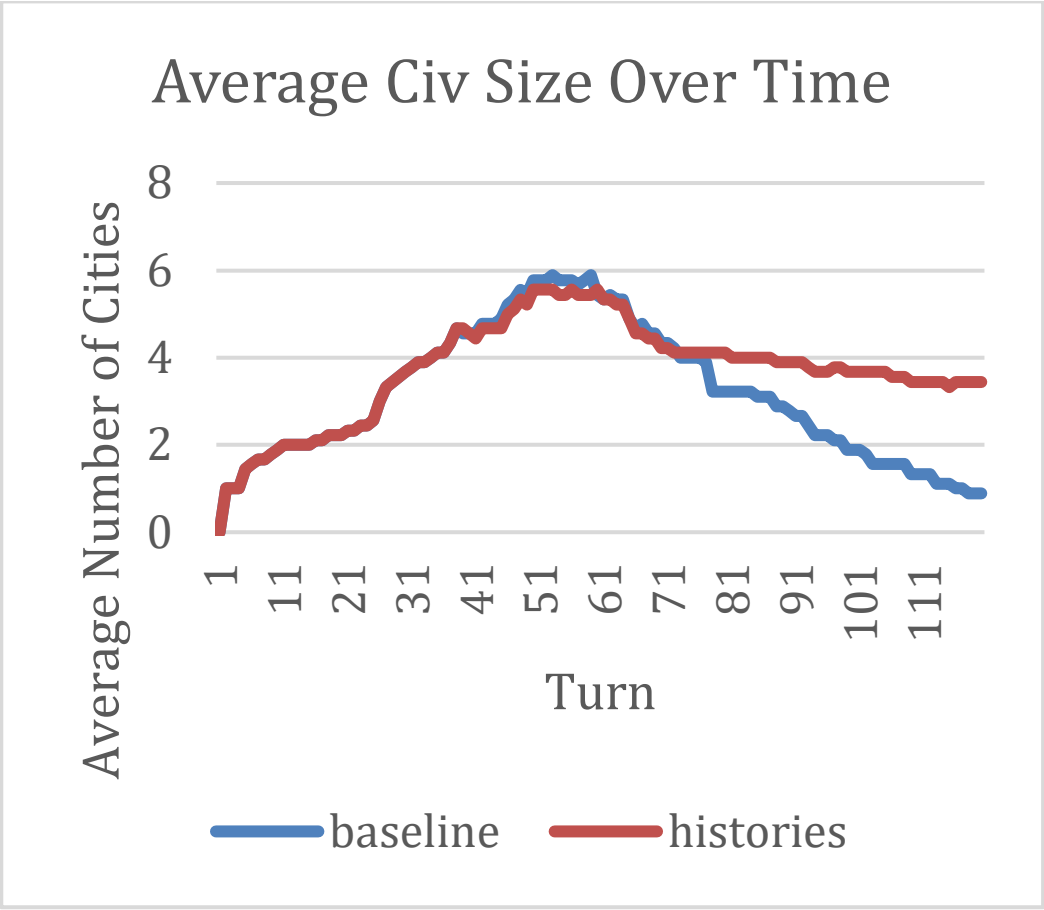

*Figure 9*: Average number of cities across ten games for baseline and histories conditions.

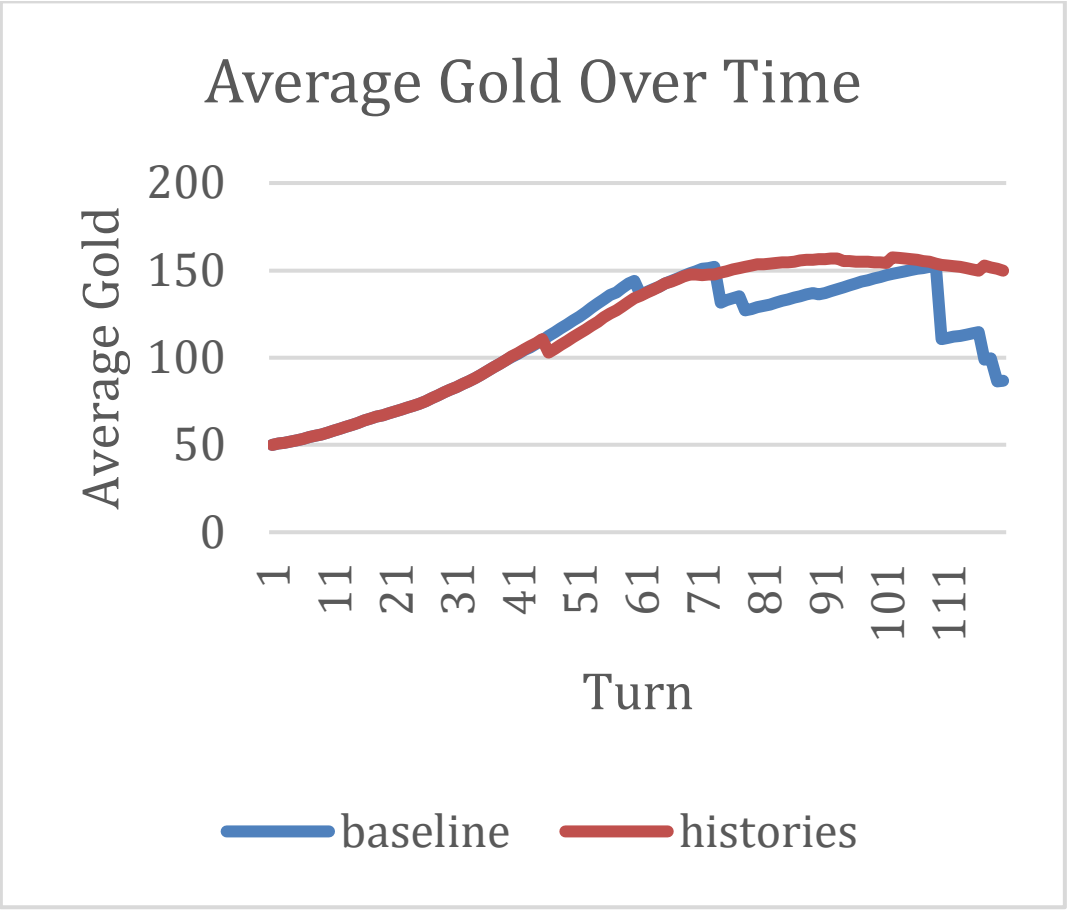

*Figure 10*: Average amount of gold across ten games for baseline and histories conditions.

Next, we analyze statistics for the cases generated in each experimental condition. In the baseline condition, more episodes are generated, but the average number of facts is smaller. In the experimental condition, there are fewer episodes to learn from, but each has more facts on average (*Table 1*). Despite having fewer examples to learn from, the histories condition significantly outperforms the baseline due to its ability to capture events at the right spatiotemporal grain size.

*Table 1*: Statistics for cases collected from the baseline and histories conditions. The histories condition significantly outperformed the baseline, even though fewer cases were generated. The average number of facts for cases in the histories condition is higher, indicating more knowledge is captured.

| Condition | Episodes | Number of Facts |
|---|---|---|
| Baseline | 164 | 64.5 |
| Histories | 117 | 76.8 |

To summarize, we have described an experiment that uses our novel qualitative spatiotemporal representation to describe battle events in Freeciv. The agent learns a model from collected cases that improves decision making for achieving successful defensive states in the game. Learning is failure-based, where the agent uses prior experience to hypothesize the outcome of a hypothetical battle given its current situation. The learned analogical model is tied to primitive actions in the domain, which suggest both achieve goals, e.g. building defensive city walls, or qualitative maximization goals, e.g. maximize the number of combat units in a region. The model learns limit points for maximization and minimization goals. We show that an agent that uses history-based representations improves its gameplay over a baseline agent that does not use extended spatiotemporal representations. A reason that the baseline condition performs poorly is that lack of extended spatiotemporal representations impedes the agent's ability to learn limit points.

## 5. Related Work

Wintermute et al. (2007) used the SOAR cognitive architecture (Laird 2019) to develop a model (SORTS) that performed various tasks within a real-time strategy game engine. SORTS also leveraged perceptual spatial representations to guide decision making but focused on models of attention rather than episodic representation. Jones and Laird (2019) laid out a general framework for event cognition within the Common Model of Cognition, which is a theoretical proposal of the computational processes underlying cognition (Laird, Lebiere, and Rosenbloom 2017). This framework focused on architectural specifications within cognitive architectures, and not the specifics of event representation. Jones (2022) provides a comprehensive implementation of these ideas in the SOAR cognitive architecture but did not leverage extended spatiotemporal representations.

## 6. Discussion and Future Work

We have described an experiment in which our agent learns qualitative states of defense in a strategy game from recording spatiotemporal traces of military battles. Empirical evidence from this experiment shows how perception of space affects event individuation and temporal segmentation. We show that history-based representations improve learning because they segment episodes based on change in the environment, and capture knowledge at a useful spatiotemporal grain size.

The event representations developed in this work can be construed as a framework for grounding quantity-conditioned processes in perceptual phenomena. In general, the spatial perception of these events affects how they persist, and therefore their duration. This applies to many kinds of events that we encounter in our everyday lives, e.g. finishing my morning cup of coffee, of class being over, etc. Often the relevant spatial bounds can be defined by containers, i.e. mugs and classrooms. In these cases, incorrect perception of the container leads to degenerate conclusions. If instead one considers the coffee shop as the container, then it is likely that a whole day goes by without finishing a cup of coffee. Here, the quantity condition (no more liquid coffee) is not met, because there is likely always coffee somewhere in the cafe. Containers are not always so readily available, however, as in the instance of representation the spatial extent of military battles. In fact, Hayes, thought that there were likely many kinds of histories that aligned with out commonsense notions of the world (Hayes, 1995). For the same reason that unbounded representations of change are likely inadequate for certain kinds of phenomenon (hence the ontology for liquids), histories are useful for event perception for continuous entities. Spilled milk, oil spills, spreading butter, and swarms are all phenomena for which the kinds of representations used in the work might help with episodic learning.

## Acknowledgements

This research was sponsored by the US Air Force Office of Scientific Research under award number FA95550-20-1- 0091.

## References

Allen, J. F. (1983). Maintaining knowledge about temporal intervals. Communications of the ACM, 26(11), 832–843.

Cohn, A. G., Bennett, B., Gooday, J., & Gotts, N. M. (1997). Qualitative Spatial Representation and Reasoning with the Region Connection Calculus. GeoInformatica, 1(3), 275–316.

Forbus, K. D. (2019). Qualitative Representations: How People Reason and Learn about the Continuous World. MIT Press.

Forbus, K. D., Ferguson, R. W., Lovett, A., & Gentner, D. (2017). Extending SME to Handle Large-Scale Cognitive Modeling. Cognitive Science, 41(5), 1152–1201.

Forbus, K. D., Gentner, D., & Law, K. (1995). MAC/FAC: A model of similarity-based retrieval. Cognitive Science, 19(2), 141–205.

Forbus, K. D., & Hinrichs, T. (2019). Qualitative Reasoning about Investment Decisions. 7.

Hancock, W., & Forbus, K. D. (2021). Qualitative Spatiotemporal Representations of Episodic Memory for Strategic Reasoning.

Hancock, W., Forbus, K. D., & Hinrichs, T. R. (2020). Towards Qualitative Spatiotemporal Representations for Episodic Memory. 33rd International Workshop on Qualitative Reasoning.

Hayes, P. J. (1979). The Naive Physics Manifesto. Expert Systems in the Microelectronic Age.

Hayes, P. J. (1989). Naive Physics I: Ontology for Liquids. In Readings in qualitative reasoning about physical systems (pp. 484–502). Morgan Kaufmann Publishers Inc.

Hayes, P. J. (1995). The Second Naive Physics Manifesto.

Hinrichs, T. R., & Forbus, K. D. (2016). Qualitative Models for Strategic Planning. Proceedings of the Third Annual Conference on Advances in Cognitive Systems, 18.

Jones, S. J., & Laird, J. (2019). Anticipatory Thinking in Cognitive Architectures with Event Cognition Mechanisms. COGSAT@AAAI Fall Symposium.

Laird, J. E. (2019). The Soar Cognitive Architecture. MIT Press.

Laird, J. E., Lebiere, C., & Rosenbloom, P. S. (2017). A Standard Model of the Mind: Toward a Common Computational Framework across Artificial Intelligence, Cognitive Science, Neuroscience, and Robotics. AI Magazine, 38(4), Article 4.

McLure, M. D., Friedman, S. E., & Forbus, K. D. (2015). Extending Analogical Generalization with Near-Misses. Twenty-Ninth AAAI Conference on Artificial Intelligence. Twenty-Ninth AAAI Conference on Artificial Intelligence.

Mohan, S., Klenk, M., Shreve, M., Evans, K., Ang, A., & Maxwell, J. (2020). Characterizing an Analogical Concept Memory for Architectures Implementing the Common Model of Cognition. *Advances in Cognitive Systems*.

Wintermute, S., Xu, J., & Laird, J. (2007). SORTS: A Human-Level Approach to Real-Time Strategy AI. Proceedings of the AAAI Conference on Artificial Intelligence and Interactive Digital Entertainment, 3(1), Article 1.

Zacks, J. M. (2020). Event Perception and Memory. Annual Review of Psychology, 71, 165–191.